\documentclass{article}

\usepackage{arxiv}
\usepackage{amsmath, amsfonts, amssymb}
\usepackage{graphicx}
\usepackage{hyperref}
\usepackage{booktabs}
\usepackage{float}
\usepackage{tikz}
\usetikzlibrary{arrows, positioning}
\makeatletter
\def\headeright{}
\def\undertitle{}
\makeatother

\title{From Agent Loops to Deterministic Graphs:\\
Execution Lineage for Reproducible AI-Native Work}

\author{
  Josh Rosen \\
  ThruWire, Inc.
  \And
  Seth Rosen \\
  ThruWire, Inc.
}

\begin{document}

\maketitle

\begin{abstract}
Large language model systems are increasingly deployed as agentic workflows that interleave reasoning, tool use, memory, and iterative refinement. These systems are effective at producing answers, but they often rely on implicit conversational state, making it difficult to preserve stable work products, isolate irrelevant updates, or propagate changes through intermediate artifacts.

We introduce execution lineage: an execution model in which AI-native work is represented as a directed acyclic graph (DAG) of artifact-producing computations with explicit dependencies, stable intermediate boundaries, and identity-based replay. The goal is not to make the model a better one-shot writer, but to make evolving AI-generated work maintainable under change.

We compare execution-lineage replay against loop-centric update baselines on two controlled policy-memo update tasks. In an unrelated-branch update, DAG replay preserved the final memo exactly in all runs, with zero churn and zero unrelated-branch contamination, while loop baselines regenerated the memo and frequently imported unrelated context. In an intermediate-artifact edit, all systems reflected the new constraint in the final memo, but only DAG replay achieved perfect upstream preservation, downstream propagation, unaffected-artifact preservation, and cross-artifact consistency.

These results show that final answer quality and maintained-state quality are distinct. Strong loop baselines can remain competitive at producing polished final outputs when the task is a bounded synthesis/update problem and all current sources fit in context, but immediate task success can mask partial state inconsistency that may compound over future revisions. Execution lineage provides stronger guarantees about what should change, what should remain stable, and how work evolves across revisions.
\end{abstract}

\section{Introduction}

Agent-based LLM systems have emerged as a dominant paradigm for applying large language models to complex tasks. Approaches such as ReAct \cite{yao2023react} combine reasoning and action within iterative loops, enabling dynamic tool usage and adaptive planning.

This paradigm has been productive because it matches the interface of current language models: a prompt is assembled, the model responds, tools may be invoked, and the loop repeats until an answer appears satisfactory. In short-horizon tasks this is often sufficient. A user can tolerate some variance, intermediate reasoning can remain transient, and restarting the loop is usually cheaper than building explicit structure.

In practice, such systems rarely expose the raw model alone. They run inside an \emph{agent harness}: a surrounding control layer that assembles prompts, manages tools and memory, and decides when execution should continue, branch, or terminate. Recent research has started to formalize the harness itself as an object of study rather than a hidden implementation detail \cite{pan2026nlah,zhang2025harness}. This shift is important because many system behaviors traditionally attributed to the model are in fact produced by the harness.

The problem changes once the work becomes multi-step, stateful, and long-lived. In those settings, the system is no longer producing a single answer in one conversational thread. It is generating intermediate research, analyses, transformations, and reports that may be revised later by humans or reused by other agents. What matters is not only whether the final answer is plausible, but whether the system can explain what it depended on, selectively recompute what changed, and preserve stable boundaries between stages of work.

Despite their flexibility, these systems exhibit systemic limitations:

\begin{itemize}
\item \textbf{Non-deterministic execution}: identical inputs may produce different outputs.
\item \textbf{Implicit dependencies}: execution structure is embedded in conversational history.
\item \textbf{Limited reproducibility}: intermediate reasoning is not preserved in structured form.
\item \textbf{Global recomputation}: small changes require full re-execution.
\end{itemize}

These properties are acceptable for single-task workflows but break down for long-lived systems where consistency and incremental updates are required.

We propose a shift from \textbf{prompt lineage}---where state is encoded in evolving prompts---to \textbf{execution lineage}, where computation is represented explicitly as a graph.

The distinction is architectural. Prompt lineage records how prompts and transcripts evolved over time, but it does not provide a stable substrate on which downstream computation can depend. Execution lineage instead externalizes the structure of the work itself: each unit of work declares its inputs, its dependencies, its output contract, and its execution identity. Under this model, intermediate artifacts are first-class objects rather than incidental text inside a prompt window.

Our research is motivated by a practical observation: many failures attributed to ``LLM unreliability'' are in fact failures of system structure. If upstream results are passed around as ad hoc context, then reuse is heuristic, inspection is manual, and re-running a workflow means reconstructing state through prompts. If the same work is represented as a graph of explicit execution units with stable intermediate boundaries, then replay, observability, and partial recomputation become system properties rather than prompt engineering tricks.

A central distinction in this paper is between deliverable quality and maintained-state quality. A loop may produce a polished final answer from the current source bundle, yet still fail to preserve unaffected work, isolate unrelated updates, or maintain consistency across intermediate artifacts. That failure may not be visible after a single update, but it can leave behind partially incoherent state for future revisions. Execution lineage targets this second class of quality.

This paper makes three contributions. First, we articulate execution lineage as a computational abstraction for AI-native workflows. Second, we formalize a DAG-based execution model that separates authored structure from model-time generation and gives intermediate artifacts stable execution identities. Third, we provide a controlled empirical comparison showing that the advantage of execution lineage is clearest on maintained-state quality rather than final prose quality: DAG replay preserves stable outputs under unrelated updates and propagates intermediate artifact edits with stronger cross-artifact consistency than loop-centric baselines.

The paper's scope is intentionally narrow. Our primary concern is \emph{how} AI-native work executes: what the unit of recomputation is, what makes a prior result replayable, and how dependency changes propagate. Richer questions about the semantics of intermediate artifacts or about multi-party co-authoring are important, but they are orthogonal to the execution-lineage thesis developed here.

\section{Related Work}

\subsection{Surveys and Taxonomies}

The literature on LLM agents has already grown large enough to generate several dedicated surveys. Recent reviews cover planning \cite{huang2024planning}, memory \cite{zhang2024memorysurvey}, broader workflow paradigms spanning tool use, planning, and feedback learning \cite{li2024review}, and, more recently, memory mechanisms and evaluation through early 2026 \cite{du2026memorysurvey}. Together they show that the field has largely converged on the agent as a composite system rather than a raw model invocation. Most survey taxonomies, however, still classify systems by capabilities such as planning, reflection, memory, and tool use. Our taxonomy cuts across those dimensions by asking whether intermediate work remains embedded in prompts and transcripts or is exposed as stable computational structure with declared dependencies and replayable boundaries.

\subsection{Agentic LLM Systems}

ReAct \cite{yao2023react} introduced interleaved reasoning and action. Toolformer \cite{schick2023toolformer} trains models to use tools via self-supervision. A parallel line of work studies agent frameworks and multi-agent organizations, including CAMEL \cite{li2023camel}, AutoGen \cite{wu2023autogen}, MetaGPT \cite{hong2023metagpt}, and Voyager \cite{wang2023voyager}. These systems show that substantial capability gains can be achieved by placing the model inside richer interaction loops, role structures, and tool environments.

These approaches share a control-loop architecture:
\begin{equation}
s_{t+1} = f(s_t, \text{LLM}, \text{tools})
\end{equation}
where state is implicit.

The strength of this family of systems is adaptability: loops can choose tools, revise prompts, and redirect work at runtime, which is especially useful for short-horizon and open-ended tasks. However, the same flexibility creates structural ambiguity. Dependencies are often encoded in conversational state rather than explicit program structure, and two executions that are semantically ``the same'' can differ because context was assembled differently or reasoning unfolded in another order. Recent work on planning agents, including Tree-of-Thoughts \cite{yao2023tree}, improves search over reasoning trajectories but retains this basic property. In our framing, these methods improve inference \emph{inside} a run, whereas execution lineage addresses structure \emph{across} runs.

\subsection{Agent Harnesses and Runtimes}

The term \emph{agent harness} has recently gained research traction as a way to name the surrounding machinery that makes an LLM agent operational. Natural-Language Agent Harnesses \cite{pan2026nlah} explicitly studies harness engineering as a portable object, arguing that important agent behavior lives in editable control logic rather than only in model weights or prompts. General Modular Harness for LLM Agents in Multi-Turn Gaming Environments \cite{zhang2025harness} similarly studies harness design as a composition of modules such as perception, memory, and reasoning.

This line of work is closely related to ours because it externalizes the wrapper around the model. Memory-oriented work pushes in the same direction: Agent Workflow Memory \cite{wang2024awm}, On the Structural Memory of LLM Agents \cite{zeng2024structuralmemory}, WorkflowLLM \cite{fan2024workflowllm}, and more recent work such as LEGOMem \cite{han2025legomem}, Memory-R1 \cite{yan2025memoryr1}, and Memori \cite{borro2026memori} all treat persistent procedural state as a meaningful systems layer rather than mere conversational carry-forward.

Our contribution differs in emphasis. Harness and memory papers ask how agent behavior can be encoded, adapted, or transferred at the controller level. We ask what execution model the harness should implement when workflows need deterministic replay, explicit lineage, and partial recomputation. In that sense, execution lineage can be understood as a systems-level proposal for what a rigorous harness runtime should optimize around. It is compatible with memory-augmented agents, but it is not reducible to memory. A memory system decides what to retain and retrieve; an execution-lineage system decides what the unit of computation is, what its dependencies are, and what exactly must be rerun when some upstream state changes.

\subsection{Reasoning Traces, Search, and Intermediate Steps}

Work on chain-of-thought and related prompting methods establishes an important empirical premise for our paper: intermediate reasoning steps often improve final-task performance. Chain-of-Thought \cite{wei2022cot}, Self-Consistency \cite{wang2022selfconsistency}, Least-to-Most prompting \cite{zhou2022leasttomost}, and scratchpad-style methods \cite{nye2021scratchpads} all show gains from exposing intermediate computation. Search-based extensions such as Tree-of-Thoughts \cite{yao2023tree}, Language Agent Tree Search \cite{zhou2023lats}, Reflexion \cite{shinn2023reflexion}, Self-Refine \cite{madaan2023selfrefine}, and structured reflection \cite{li2023structuredreflection} push farther by revising and exploring alternative reasoning trajectories; earlier work on workflow-guided exploration \cite{liu2018workflowguided} foreshadows the same interest in reusable action structure.

These papers matter here for two reasons. First, they show that intermediate steps are often where performance gains come from. Second, they reveal the limitation of prompt-centric intermediate state: most still represent intermediate work as textual traces tied to a specific execution episode. Our proposal builds on the empirical value of intermediate reasoning while relocating its substrate from transient text into explicit execution structure.

\subsection{Workflow and DAG Systems}

Workflow and dataflow systems such as Dryad \cite{isard2007dryad} and Spark \cite{zaharia2016spark}, as well as orchestration and analytics systems such as Airflow \cite{airflowdocs} and dbt \cite{dbtdocs}, model execution around explicit dependency graphs, scheduled runs, and lineage-aware transformations.

These systems emphasize:
\begin{itemize}
\item explicit dependencies
\item deterministic execution
\item incremental recomputation
\end{itemize}

The relevance of these systems is not superficial. They reflect a mature systems answer to a recurring problem: when work becomes multi-stage and collaborative, hidden dependencies and ephemeral outputs do not scale. Data engineering historically moved from scripts and manual handoffs toward compiled graphs, materialized artifacts, and lineage-aware recomputation because these primitives were necessary for correctness and maintainability. We view AI-native workflows as reaching a similar inflection point.

That said, classical DAG systems were not designed around stochastic model calls, evolving prompts, or semantically rich intermediate artifacts. Their nodes typically represent deterministic code transformations with well-defined data contracts. Our work therefore does not merely import DAGs into AI systems; it adapts the DAG abstraction to settings where nodes may invoke LLMs, generate structured artifacts, and depend on both model configuration and upstream execution identity.

\subsection{Programmatic LLM Systems}

Emerging work treats LLMs as programs:
\begin{itemize}
\item DSPy \cite{khattab2023dspy}
\item LMQL \cite{beurer2023lmql}
\item PAL \cite{gao2023pal}
\item Program of Thoughts \cite{chen2022pot}
\end{itemize}

These frameworks introduce valuable structure around prompting, constraints, optimization, and output shaping. Nevertheless, most remain primarily sequential or call-graph oriented rather than execution-lineage oriented. They improve how a single pipeline is programmed, but they generally do not make dependency materialization, replay identity, or partial downstream invalidation the organizing abstraction of the whole system.

\subsection{Benchmarks and Reproducible Agent Environments}

Another relevant literature studies how to evaluate agents in realistic yet reproducible environments. Benchmarks such as AgentBench \cite{liu2023agentbench}, WebArena \cite{zhou2023webarena}, VisualWebArena \cite{koh2024visualwebarena}, WorkArena \cite{drouin2024workarena}, AndroidWorld \cite{rawles2024androidworld}, OSWorld \cite{xie2024osworld}, AppWorld \cite{trivedi2024appworld}, GAIA \cite{mialon2023gaia}, and LifelongAgentBench \cite{zheng2025lifelongagentbench} increasingly evaluate agents in stateful, tool-using settings rather than static prompts.

Recent 2025--2026 benchmark work sharpens the memory and persistence angle in particular. MemoryAgentBench \cite{hu2025memoryagentbench}, Evo-Memory \cite{wei2025evomemory}, and Mem2ActBench \cite{shen2026mem2actbench} all argue that static one-shot evaluation misses the harder problem of incremental accumulation, selective forgetting, and task-conditioned reuse. These benchmarks are not execution-lineage systems, but they are strong evidence that the field is moving toward persistent, longitudinal evaluation rather than isolated agent episodes.

This benchmark literature is important because it increasingly treats \emph{environmental reproducibility} and \emph{execution-based evaluation} as first-class concerns. However, benchmark reproducibility is not workflow reproducibility. A benchmark may make the external environment resettable and measurable while still leaving the internal execution structure of the agent implicit. Our work is complementary: we focus on making the agent's own computational lineage explicit and incrementally reusable.

\subsection{Workflow Provenance and Interactive Analysis}

There is also an emerging line of work on provenance-aware agent systems in scientific and workflow settings. LLM Agents for Interactive Workflow Provenance \cite{souza2025provenance} studies how LLM agents can query and analyze workflow provenance records through modular interfaces. Connecting Large Language Model Agent to High Performance Computing Resource \cite{ma2025hpc} examines how tool-using agents can operate over parallel and distributed execution substrates.

These papers are adjacent to our thesis because they treat workflows and provenance as serious systems objects rather than informal prompt context. The difference is that provenance work typically assumes an existing workflow system whose traces are queried after the fact. Execution lineage, as we define it, moves provenance into the execution substrate itself: lineage helps determine identity, replay, and invalidation during execution.

\subsection{Reproducibility and Evaluation}

LLM reproducibility challenges are well documented \cite{chen2023evaluating}. Variance arises from sampling, model updates, and tool interactions.

Prior work has correctly emphasized that reproducibility in language model systems is limited by multiple factors, including decoding stochasticity, external API variation, and infrastructure drift. Work on interactive evaluation \cite{abramson2022multimodal} and realistic agent benchmarks \cite{rawles2024androidworld,xie2024osworld} likewise shows that even controlled environments exhibit sensitivity to rollout details and test conditions. We do not claim to eliminate all sources of variance. Instead, we isolate a source that is architectural and tractable: execution structure itself. Even when model and tool behavior are held fixed, prompt-oriented systems often reconstitute state heuristically at runtime. This introduces unnecessary instability and obscures provenance.

Our work complements reproducibility research by shifting attention from model behavior alone to the execution substrate around the model. The question is not only whether a model can reproduce a token sequence, but whether the system can reproduce a computational step, restore its exact intermediate state, and determine precisely which downstream work should be invalidated when an upstream artifact changes.

\section{Execution Lineage Model}

We define a workflow as:
\begin{equation}
G = (V, E)
\end{equation}

where $V$ is a set of executable nodes and $E$ is a set of directed dependency edges. Each edge $(u, v) \in E$ states that node $v$ may consume only artifacts explicitly produced by $u$ and declared as part of its input surface.

Each node:
\begin{equation}
v = (\mathcal{I}_v, f_v, \mathcal{O}_v)
\end{equation}

where $\mathcal{I}_v$ denotes the resolved local input state available to node $v$, $f_v$ is the node-local execution procedure, and $\mathcal{O}_v$ is the structured output artifact selected as the node's result. Artifacts are not arbitrary transcripts; they are typed outputs with stable boundaries intended for downstream consumption.

This formulation differs from agent loops in two ways. First, the admissible context for each node is declared before execution rather than reconstructed opportunistically from conversational state. Second, node outputs persist as addressable artifacts with explicit lineage, enabling downstream reuse and inspection.

\subsection{Design Principles}

The execution-lineage model can be summarized by four design principles.

\paragraph{Explicit dependency declaration.}
A node should not be allowed to consume undeclared upstream state. This principle forces the graph to carry the causal structure of the workflow rather than hiding it in prompt assembly logic.

\paragraph{Local visibility boundaries.}
Each node should see only the context and dependency artifacts required for its task. Besides improving provenance, this reduces accidental coupling, lowers prompt bloat, and makes it easier to understand why a node produced a given output.

\paragraph{Identity-based reuse.}
Reuse should be based on proof of equivalence, not similarity of prompts or approximate resemblance of outputs. A runtime should be able to say not merely that two executions ``look close enough,'' but that they share the same declared structure and resolved inputs.

\paragraph{Deterministic publication.}
Even when execution involves internal iteration, branching, or validation, the boundary exposed downstream should be stable and canonical. Downstream nodes should depend on a published result, not on whatever happened to be most recently present in a conversational trace.

Taken together, these principles describe a runtime whose main object is not the prompt or the transcript, but the dependency-respecting publication of intermediate computation.

\subsection{Intermediate State and Typed Local Boundaries}

The core execution object is the intermediate state boundary. In our implementation this is materialized as an artifact, but the deeper claim is executional rather than representational: the system needs a stable, inspectable unit that downstream computation can depend on. An artifact may contain free-form model output, but it is treated by the runtime as a structured unit with an identity, type, and provenance. This matters because the role of an intermediate result is not merely to be read by a human; it is to become a stable dependency surface for later computation.

Each node executes against a typed local state consisting of three classes of entries: immutable context provided at invocation time, immutable dependency artifacts materialized from upstream nodes, and mutable local artifacts produced during the node's own execution. This separation is important. Without it, a runtime cannot distinguish inherited state from newly produced state, and downstream reasoning about provenance collapses back into transcript reconstruction.

\subsection{Dependency Materialization}

Dependencies are resolved against actual inputs rather than only logical node names. Concretely, an upstream dependency is identified not only by which node produced it, but by the particular resolved input surface under which that node was executed. This prevents two semantically distinct executions of the same logical node from being aliased together.

Formally, let $k_v$ denote the execution identity of node $v$. We define:
\begin{equation}
k_v = h(\sigma_v, x_v, \{k_u : u \in \mathrm{pred}(v)\})
\end{equation}
where $\sigma_v$ is the structural specification of node $v$, $x_v$ is its resolved input hash, and $\mathrm{pred}(v)$ are its immediate predecessors. The hash function $h$ need not be cryptographically special for our argument; what matters is that node identity is a deterministic function of structure, inputs, and upstream identities.

This execution identity provides the basis for cache reuse, replay, and selective invalidation. If $k_v$ is unchanged, prior execution may be reused exactly. If it changes, downstream nodes that depend on $k_v$ are invalidated and re-executed, while unrelated branches remain untouched.

\subsection{Execution Semantics}

Execution is a forward pass:
\begin{equation}
\mathcal{O}_v = f_v(\mathcal{I}_v)
\end{equation}

Nodes execute once dependencies are satisfied. At runtime, readiness is decided from the graph rather than inferred by the model. This allows the system to schedule independent branches concurrently while preserving sequential semantics within each node's local procedure.

In practice, a node may itself contain multiple internal steps, such as context selection, model invocation, validation, or rendering. We treat these as part of $f_v$. They may be sophisticated, but they remain subordinate to a platform-controlled execution boundary. The model generates content within the node; it does not decide which nodes exist or which undeclared upstream state may be read.

\subsection{Determinism}

\textbf{Definition:}
Execution is deterministic if outputs are reproducible under:
\begin{itemize}
\item fixed inputs
\item fixed model + parameters
\item fixed tool outputs
\end{itemize}

Under these conditions, determinism means more than ``similar answers.'' It means that the system can either replay a prior node execution exactly or prove that some aspect of its identity changed and therefore recomputation is required. Determinism is therefore a property of the execution substrate, not a promise that the model itself is universally stable under all deployments.

\subsection{Replay and Partial Recomputation}

Execution lineage makes replay a first-class runtime mode. When a node's execution identity matches a previously persisted result, the runtime can restore its artifact and associated execution record rather than regenerate it heuristically. This differs from approximate caching: replay is identity-based and therefore explainable.

Similarly, partial recomputation follows directly from graph structure. If an upstream research artifact changes, only descendants whose execution identities depend on that artifact must be recomputed. The runtime can leave unrelated branches intact. In contrast, prompt-centric systems often collapse all prior work into one evolving context window, making a local change expensive because the system lacks precise invalidation boundaries.

\subsection{Invalidation Semantics}

The usefulness of an execution graph depends not only on replay, but on \emph{correct} replay refusal. A lineage-aware runtime must be able to explain both why prior work can be reused and why it can no longer be reused. Invalidation therefore becomes a semantic operation over dependency identities rather than a vague sense that ``the prompt changed.''

Consider a simple three-stage workflow consisting of retrieval, analysis, and synthesis. If only the synthesis prompt changes, the system should preserve retrieval and analysis while recomputing only synthesis. If a retrieval source changes, analysis and synthesis must be invalidated because their dependency identities are downstream of the changed source. This style of invalidation is routine in build systems and dataflow engines, but remains under-specified in many agent systems, where upstream changes are usually handled by restarting or partially reconstructing a prompt transcript.

This distinction matters operationally. Without explicit invalidation semantics, teams tend to choose between two unsatisfactory behaviors: stale reuse of outputs that no longer match current assumptions, or expensive global reruns that throw away valid prior work. Execution lineage aims to replace that ambiguity with a principled middle ground.

\subsection{Canonical Outputs and Observability}

An execution system also needs a notion of which output is authoritative at a boundary. In loop-based systems, candidate drafts, self-critiques, tool observations, and final responses are often mixed together in one transcript. A human can often infer which part ``counts,'' but the runtime itself lacks a reliable canonical object for downstream consumers.

Execution-lineage systems instead select a canonical output at each node boundary. That output may have been produced through multiple internal steps, validations, or revisions, but once selected it becomes the stable downstream dependency surface. This improves observability in two ways. First, it lets operators inspect the exact artifact that downstream nodes consumed, rather than rereading an entire transcript. Second, it creates a durable execution history in which intermediate candidates, final selected outputs, and dependency identities can be distinguished rather than conflated.

\subsection{Execution Lineage vs. Prompt Lineage}

Prompt lineage records the history of prompts, messages, and tool observations used during an interaction. It is useful for audit and narration, but it is a poor substrate for composition because it lacks stable execution boundaries. Execution lineage, by contrast, records the graph of artifact-producing computations and the identities that connect them.

The distinction parallels the difference between logs and program structure. Logs are valuable for understanding what happened after the fact; they are not the mechanism by which a system should declare dependencies or reuse work. In our model, reasoning traces may still exist, but they are auxiliary observability surfaces rather than the system's computational backbone.

\section{Execution vs Agent Loop}

\begin{figure}[H]
\centering
\begin{tikzpicture}[
  font=\small,
  scale=0.85,
  transform shape,
  node distance=7mm and 10mm,
  box/.style={draw, rounded corners, align=center, minimum width=28mm, minimum height=8mm},
  artifact/.style={draw, rounded corners, align=center, minimum width=27mm, minimum height=8mm, fill=black!3},
  note/.style={align=left, font=\footnotesize},
  line/.style={->, >=stealth, thick},
  muted/.style={draw=black!55, text=black!70}
]
\node[font=\bfseries] (looptitle) at (0,0) {Agent loop};
\node[box] (goal) at (0,-0.8) {Input / goal};
\node[box] (prompt) at (0,-2.0) {Prompt assembly};
\node[box] (runtime) at (0,-3.3) {LLM / tool loop};
\node[box] (final) at (0,-4.6) {Final answer};
\node[note] (implicit) at (3.0,-3.3) {implicit memory\\transcript state\\tool observations};
\draw[line] (goal) -- (prompt);
\draw[line] (prompt) -- (runtime);
\draw[line] (runtime) -- (final);
\draw[line] (runtime.east) .. controls +(0.8,0.75) and +(0.8,-0.75) .. (runtime.east);
\draw[line] (implicit.west) -- (runtime.east);

\node[font=\bfseries] (graphtitle) at (10.0,0) {Execution lineage};
\node[artifact] (edit) at (10.0,-0.8) {Input / edit event};
\node[artifact] (a) at (10.0,-2.0) {A: source / context\\artifact};
\node[artifact] (b) at (10.0,-3.2) {B: analysis\\artifact};
\node[artifact] (c) at (10.0,-4.4) {C: criteria / decision\\artifact};
\node[artifact] (d) at (10.0,-5.6) {D: final artifact};
\node[artifact, muted] (branchsrc) at (14.8,-2.0) {Adjacent\\branch};
\node[artifact, muted] (branchnote) at (14.8,-3.2) {side artifact};
\node[note] (recompute) at (12.8,-4.2) {edit propagates\\downstream};
\node[note] (preserve) at (14.8,-5.25) {not upstream of\\final artifact:\\preserve exactly};
\draw[line] (edit) -- (a);
\draw[line] (a) -- (b);
\draw[line] (b) -- (c);
\draw[line] (c) -- (d);
\draw[line, dashed] (branchsrc) -- (branchnote);
\draw[line, dashed] (c.east) -- (recompute.west);
\draw[line, dashed] (branchnote.south) -- (preserve.north);
\node[note] at (7.0,-5.9) {replay / preserve upstream};
\end{tikzpicture}
\caption{Agent loops carry work forward through prompt context and transcript state. Execution lineage represents work as explicit artifact-producing nodes with declared dependencies, allowing the runtime to decide which artifacts to replay, recompute, or preserve after an edit.}
\label{fig:loop-vs-dag}
\end{figure}
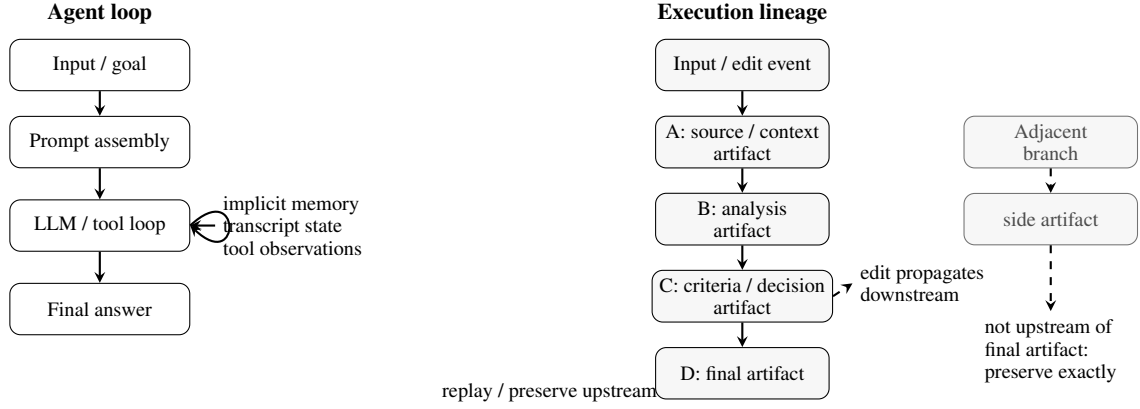

\section{Research Questions}

\begin{itemize}
\item \textbf{RQ1: Dependency isolation.} When an update occurs outside the dependency path of a final artifact, can the system preserve the final artifact exactly rather than rewriting it?
\item \textbf{RQ2: Artifact-level propagation.} When an intermediate artifact is edited, can the system update downstream dependents while preserving upstream and unrelated artifacts?
\item \textbf{RQ3: Maintained-state quality.} Do execution-lineage workflows outperform loop-centric workflows on maintained-state properties such as preservation, propagation, unrelated-branch isolation, and cross-artifact consistency?
\item \textbf{RQ4: Deliverable-quality boundary.} When all systems can update the visible final memo, do maintained-state metrics reveal differences that final-answer metrics miss?
\item \textbf{RQ5: Efficiency tradeoffs.} How do token use, model-call count, and wall-clock time differ between loop updates and DAG replay under unrelated-branch versus downstream-propagating edits?
\end{itemize}

The first two questions correspond to the two experimental interventions; the latter three characterize the broader distinction between visible deliverable correctness and maintained-state correctness.

\section{Methods}

\subsection{Study Design}

We use a within-task comparative design over two controlled policy-memo update tasks. We instantiate the tasks in a telehealth expansion scenario so that source updates, recommendation criteria, and final deliverables can be inspected concretely. Each task is run under three conditions: a naturalistic loop update, a loop update with explicit edit-event awareness, and execution-lineage DAG replay. All conditions use the same model family and source materials. The comparison is system-level rather than equal-prompt: the DAG condition receives runtime-derived dependency and edit-lineage state because that state is the architectural substrate under evaluation. Loop conditions rely on implicit reconstruction from the prior memo, current materials, and, in one condition, the edit event.

\subsection{Experimental Conditions}

For readability, we refer to these conditions in the rest of the paper as \emph{loop final update}, \emph{loop + edit event}, and \emph{DAG replay}.

\begin{itemize}
\item \textbf{Loop final update} (\texttt{loop\_real\_world\_final\_update}). A naturalistic loop baseline. It receives the prior final memo, current source materials, the requested output format, and a normal update request. It does not receive a dependency graph, affected or unaffected claim IDs, recomputation scope, allowed or disallowed artifact lists, oracle labels, or the correct answer.
\item \textbf{Loop + edit event} (\texttt{loop\_real\_world\_with\_edit\_event}). The same loop baseline, but also given the source or artifact edit event. Its purpose is to test whether the DAG advantage is merely knowing what changed. It still does not receive a dependency graph, downstream propagation plan, affected or unaffected labels, recomputation stages, oracle labels, or the correct answer.
\item \textbf{DAG replay} (\texttt{simple\_dag\_replay\_selective\_recompute}). The execution-lineage condition. It receives the artifact or source edit event, artifact identities, the dependency graph, a dependency-derived recomputation plan, and preserved artifact state. It does not receive judge-only labels, the correct final recommendation, the hidden scoring rubric, or oracle affected-claim IDs unless the runtime itself derives them.
\end{itemize}

\subsection{Tasks}

\begin{itemize}
\item \textbf{\texttt{unrelated\_branch\_noop\_update}}. This task tests dependency isolation. The maintained work product is a telehealth policy memo, while the update occurs in a provider recruiting and staffing branch that is plausible and adjacent but not upstream of the final telehealth memo. Correct behavior is to preserve the final telehealth memo exactly and avoid importing recruiting or staffing evidence into the telehealth recommendation.
\item \textbf{\texttt{intermediate\_artifact\_edit}}. This task tests artifact-level propagation. A human or agent edits an intermediate recommendation-criteria artifact to impose a new constraint: year-one telehealth expansion must be budget-neutral and must include utilization controls before chronic-care follow-up expansion can scale. Correct behavior is to preserve upstream evidence artifacts, propagate the edit downstream to the implementation plan and final memo, preserve unaffected artifacts, and keep the final memo consistent with the edited criteria.
\end{itemize}

\subsection{Metrics}

\paragraph{Unrelated-branch update metrics}
\begin{itemize}
\item exact preservation, measured by \texttt{final\_output\_exact\_match} and \texttt{final\_output\_hash\_preserved}
\item stable-artifact preservation and churn, including \texttt{stable\_artifact\_hash\_preservation} and \texttt{unnecessary\_churn\_rate}
\item unrelated-branch contamination, measured by \texttt{unrelated\_branch\_contamination\_rate}
\item semantic correctness proxies, including \texttt{output\_faithfulness\_score} and \texttt{current\_state\_precision\_score}
\item efficiency measures: input and output tokens, model-call count, and wall-clock time
\end{itemize}

\paragraph{Intermediate-artifact edit metrics}
\begin{itemize}
\item final-memo constraint reflection, measured by \texttt{final\_memo\_constraint\_reflection}
\item cross-artifact consistency, measured by \texttt{cross\_artifact\_consistency\_score}
\item stable-artifact preservation and upstream churn, including \texttt{stable\_artifact\_hash\_preservation}, \texttt{upstream\_churn\_rate}, and \texttt{unaffected\_artifact\_preservation}
\item downstream propagation, measured by \texttt{downstream\_propagation\_recall}
\item semantic correctness proxies, including \texttt{output\_faithfulness\_score} and \texttt{current\_state\_precision\_score}
\item efficiency measures: input and output tokens, model-call count, and wall-clock time
\end{itemize}

For the unrelated-branch task, contamination is counted only when the final memo uses, cites, or imports content from the provider-recruiting branch into the telehealth recommendation. The presence of the provider-recruiting artifact elsewhere in the DAG state is not counted as contamination. For the intermediate-edit task, cross-artifact consistency measures whether the final memo remains aligned with the maintained intermediate artifacts after the criteria edit. Constraint reflection measures whether the final memo includes the new budget-neutrality and utilization-control constraint. These semantic metrics are evaluated from stored outputs and judge inputs. Hash preservation, churn, token counts, model-call counts, and wall-clock time are computed directly from run artifacts; hash-preservation metrics compare the updated artifact against the pre-edit artifact, and a score of $1.0$ means exact byte-level preservation.

Each task-condition pair is run with $n=3$ repeats and semantic judging uses two judge passes per repeat. The reported run uses \texttt{GPT-5.2} at temperature $0.7$ with repo commit \texttt{f6cc8679ab1178b22fa01c263aa14e77454081ee}. Deterministic metrics are computed from artifact hashes, replay records, and churn heuristics, while judge-like scores are used only for specific semantic fields such as contamination, constraint reflection, and cross-artifact consistency. No generation condition receives judge-only metadata.

\section{Experimental Setup}

The controlled policy-memo setting is instantiated as a telehealth expansion memo about whether a multistate health system should expand behavioral-health and chronic-care follow-up visits. The DAG condition represents the work as staged artifacts including utilization context, reimbursement context, operations context, access/cost context, claim matrix, tension analysis, recommendation criteria, implementation plan where applicable, and final memo. The unrelated-branch task adds an adjacent provider-recruiting branch outside the final memo's dependency path. The intermediate-edit task modifies recommendation criteria with a budget-neutrality and utilization-control constraint.

Across all conditions, we hold constant:
\begin{itemize}
\item the model family
\item the source materials
\item the task domain
\item stored rendered prompts
\item stored outputs
\item stored judge inputs and outputs
\end{itemize}

The loop conditions do not receive the dependency graph. This is intentional: the comparison is between implicit context reconstruction and explicit runtime lineage, not between equalized prompt contents.

\section{Results}

\subsection{Overview: Maintained-State Quality}

The controlled update tasks show a clear separation between final deliverable correctness and maintained-state correctness. In the unrelated-branch update, DAG replay preserved the final memo exactly while the loop baselines regenerated and often contaminated it. In the intermediate-artifact edit, all systems updated the final memo correctly, but only DAG replay maintained perfect upstream preservation, downstream propagation, and cross-artifact consistency. Preservation and propagation are two sides of maintained-state quality: a system must know both when an artifact should remain fixed and when a change should flow downstream. Tables~\ref{tab:dependency-isolation} and \ref{tab:artifact-propagation} summarize the two controlled update tasks.

\begingroup
\setlength{\intextsep}{8pt}
\begin{table}[H]
\centering
\small
\refstepcounter{table}
\textbf{Table~\thetable:} Dependency-isolation results for the unrelated-branch update ($n=3$). The correct behavior is exact preservation without unrelated-branch contamination.\par\vspace{4pt}
\label{tab:dependency-isolation}
\begin{tabular}{lcccccc}
\toprule
Condition & Exact preserve & Churn & Contam. & Input tokens & Model calls & Wall-clock \\
\midrule
Loop final update & 0.00 & 0.923 & 0.667 & 11655 & 1 & 33.5s \\
Loop + edit event & 0.00 & 0.908 & 1.000 & 11694 & 1 & 55.0s \\
DAG replay & 1.00 & 0.000 & 0.000 & 382 & 1 & 7.9s \\
\bottomrule
\end{tabular}
\end{table}

\begin{table}[H]
\centering
\footnotesize
\refstepcounter{table}
\textbf{Table~\thetable:} Artifact-level propagation results for the intermediate-criteria edit ($n=3$). All systems reflected the new constraint; only DAG replay preserved upstream state and maintained full cross-artifact consistency.\par\vspace{4pt}
\label{tab:artifact-propagation}
\begin{tabular}{lcccccccc}
\toprule
Condition & \shortstack{Constraint\\reflected} & \shortstack{Cross-artifact\\consist.} & \shortstack{Stable artifact\\preservation} & \shortstack{Downstream\\propagation} & \shortstack{Upstream\\churn} & Input tokens & \shortstack{Model\\calls} & \shortstack{Wall-\\clock} \\
\midrule
Loop final update & 1.00 & 0.50 & -- & -- & -- & 11404 & 1 & 38.7s \\
Loop + edit event & 1.00 & 0.50 & -- & -- & -- & 11464 & 1 & 31.1s \\
DAG replay & 1.00 & 1.00 & 1.00 & 1.00 & 0.00 & 8461 & 2 & 73.6s \\
\bottomrule
\end{tabular}
\end{table}
\endgroup

\subsection{Dependency Isolation: Unrelated Branch Update}

The unrelated-branch task is the most direct dependency-isolation test in this study. The correct behavior is preservation: the system should determine that the recruiting update is outside the final memo's dependency path and therefore should not alter it. As Table~\ref{tab:dependency-isolation} shows, DAG replay preserved the final memo exactly in $3/3$ runs, with \texttt{final\_output\_exact\_match} and \texttt{final\_output\_hash\_preserved} both equal to $1.00$. It also achieved \texttt{stable\_artifact\_hash\_preservation} of $1.00$, zero churn, and zero unrelated-branch contamination.

The loop baselines regenerated the final memo in $3/3$ runs. The loop final update condition contaminated the memo in $2/3$ runs, while the edit-event loop contaminated it in $3/3$ runs. This matters because the edit-event loop knew what changed at the source level yet still regenerated the memo and imported unrelated branch content. In this task, source-change awareness alone was insufficient; the distinguishing factor was dependency structure rather than edit-event awareness alone.

The efficiency difference was also large on this unrelated-branch update. DAG replay used $382$ input tokens versus about $11.7$k for either loop baseline, or about $30.5\times$ fewer than loop final update and $30.6\times$ fewer than the edit-event loop. Output tokens were also lower for DAG replay ($440.7$) than for the two loops ($2189.3$ and $2078.3$). It was also about $4.2\times$ faster than loop final update ($33.5$s vs.\ $7.9$s) and about $6.9\times$ faster than the edit-event loop ($55.0$s vs.\ $7.9$s). Faithfulness and current-state precision followed the same pattern: DAG replay scored $1.0$ on both, while the loop baselines in this study scored $0.0$ faithfulness with current-state precision of $0.333$ and $0.0$, respectively. In this task, DAG replay achieved exact preservation and lower contamination with substantially lower token use and wall-clock time.

\subsection{Artifact-Level Propagation: Intermediate Criteria Edit}

The intermediate-artifact task provides the complementary downstream-propagation test and is at least as important conceptually as the unrelated-branch update. Here the correct behavior is not preservation alone. The system must propagate a meaningful edit through downstream artifacts while preserving upstream evidence and unrelated state. Table~\ref{tab:artifact-propagation} reports the core task-specific metrics.

All three conditions reflected the new constraint in the final memo, with \texttt{final\_memo\_constraint\_reflection} equal to $1.00$ in every case. All three also achieved \texttt{output\_faithfulness\_score} and \texttt{current\_state\_precision\_score} of $1.00$. A final-memo-only evaluation would therefore treat all systems as successful. The maintained-state metrics reveal the distinction.

DAG replay achieved perfect \texttt{stable\_artifact\_hash\_preservation} ($1.00$), \texttt{downstream\_propagation\_recall} ($1.00$), \texttt{upstream\_churn\_rate} ($0.00$), \texttt{unaffected\_artifact\_preservation} ($1.00$), and \texttt{cross\_artifact\_consistency\_score} ($1.00$). The loop baselines reflected the edit in the final memo, while DAG replay preserved and updated the surrounding artifact state consistently. Both loop baselines reflected the new constraint in the final memo, but each reached only $0.50$ cross-artifact consistency because the updated final deliverable was not consistently aligned with maintained intermediate state. All three conditions had \texttt{final\_output\_exact\_match} and \texttt{final\_output\_hash\_preserved} equal to $0.0$, which is expected here because the final memo should change.

This task also shows that the results do not support a general claim of wall-clock superiority. DAG replay used fewer input tokens than the loops ($8461$ versus $11404$ and $11464$, or about $1.35\times$ and $1.36\times$ less input context), but it produced more output tokens ($4311.3$ versus $2418.7$ and $1981.0$) and was slower in wall-clock time because the implementation recomputed downstream artifacts sequentially in two model calls. It took $73.6$s versus $38.7$s for loop final update and $31.1$s for the edit-event loop, making it about $1.9\times$ and $2.36\times$ slower, respectively.

\subsection{Deliverable Quality vs.\ Maintained-Work Quality}

The intermediate-edit task illustrates why final-output metrics alone are insufficient. A final-memo-only evaluation would score all three systems as successful because all reflected the new constraint. The maintained-state metrics reveal the distinction: only DAG replay preserved upstream artifacts, propagated the edit downstream, preserved unaffected artifacts, and maintained consistency between intermediate artifacts and the final memo.

This result sharpens the comparison target. Strong loop baselines can remain competitive on final deliverable quality when the task is a bounded synthesis or update problem and all current sources fit in context. The main difference in this study appears in the maintained state: preservation, isolation, propagation, and cross-artifact consistency under revision.

\subsection{Efficiency Tradeoffs}

The reported results do not support a blanket claim that DAG replay is always faster. On the unrelated-branch update, DAG replay achieved higher maintained-state correctness with lower token use and wall-clock time because it reused preserved state and scoped recomputation tightly. On the intermediate edit, DAG replay used less input context but was slower in wall-clock time due to sequential multi-stage replay. The implication is narrower: execution lineage improves maintained-state quality under change, while efficiency depends on the structure of the update and the runtime's replay implementation.

\subsection{Information Boundaries}

\begin{table}[h]
\centering
\scriptsize
\resizebox{\textwidth}{!}{%
\begin{tabular}{lcccccc}
\toprule
Condition & Prior memo/artifacts & Current sources & Edit event & Dependency graph & Selective recomputation & Oracle labels \\
\midrule
Loop final update & yes & yes & no & no & no & no \\
Loop + edit event & yes & yes & yes & no & no & no \\
DAG replay & yes & scoped & yes & yes & yes & no \\
\bottomrule
\end{tabular}
}
\caption{Information boundaries by condition. The comparison is system-level, not an equal-prompt comparison.}
\label{tab:info-boundaries}
\end{table}

\section{Discussion}

\subsection{Why This Matters}

Agent loops optimize for task completion. Execution graphs optimize for:
\begin{itemize}
\item persistence
\item evolution
\item collaboration
\end{itemize}

This difference becomes more important as AI systems move from one-shot assistants to persistent work environments. In many real settings, the goal is not merely to get an answer once. The goal is to build a body of work that can be revised, extended, audited, and shared across humans and agents. A system that stores only prompts and final answers leaves too much of the work trapped inside transient execution.

Execution lineage changes the unit of improvement. Instead of iterating only on a final output, teams can iterate on the structure that produces outputs. A change to an upstream methodology, research constraint, or transformation rule can propagate through the graph in a principled way. This resembles the transition seen in data engineering, where the object of engineering shifted from isolated scripts to the dependency system that generated downstream tables and reports.

The practical collaboration benefit follows from this, but is not the paper's main claim. Once intermediate boundaries and dependencies are explicit, humans and agents can intervene at meaningful points rather than by restating instructions into a loop. However, the novelty we emphasize here is the runtime substrate that makes such interventions precise: explicit invalidation boundaries, replayable identities, and deterministic scheduling over a graph.

\subsection{Immediate Task Success vs.\ Maintained-State Quality}

The experiments suggest that final prose quality is an incomplete measure for long-lived AI work because it rewards immediate task completion more than the health of the maintained state. In the intermediate-artifact task, all systems produced a final memo that reflected the new constraint, so a final-answer-only evaluation would treat them as successful. The maintained-state metrics reveal the difference: the loop baselines updated the visible deliverable but left only partial cross-artifact consistency, while DAG replay preserved upstream artifacts, propagated the edit downstream, and kept the final output aligned with the intermediate state.

The deeper risk is longitudinal state drift. A loop-centric update can satisfy the immediate task while leaving the surrounding artifact state only partially coherent. That may be acceptable after one revision, but repeated updates could compound the inconsistency as future work inherits artifacts that no longer agree about the current criteria, evidence base, or implementation assumptions. Our experiments expose this mechanism but do not directly measure long-horizon accumulation; testing whether these inconsistencies compound over many revisions remains future work.

\subsection{Controlled Propagation}

The no-op result should not be read narrowly as a claim that execution lineage is useful only when work should remain unchanged. It is one instance of a broader property: dependency-aware control over change propagation. The same mechanism that prevents irrelevant updates from reaching the final memo also allows intermediate artifact edits to propagate only to downstream dependents. Preservation and propagation are two sides of the same execution-lineage property.

\subsection{When Not to Regenerate}

The unrelated-branch update illustrates a failure mode of loop-centric systems that is rarely measured: unnecessary regeneration. The correct behavior was to leave the final memo untouched. Both loop baselines rewrote it, and the edit-event loop was not protected by knowing which source changed. This suggests that source-change awareness is not enough; a system also needs dependency structure to decide whether a change is relevant to a given artifact.

\subsection{Where Execution Graphs Help Most}

Execution lineage is unlikely to matter equally for all AI tasks. Its advantages should be most visible in workflows with several properties.

\begin{itemize}
\item \textbf{Dependency isolation}: the workflow contains branches whose outputs should remain stable under adjacent updates.
\item \textbf{Unrelated-branch isolation}: plausible nearby edits should not contaminate downstream artifacts.
\item \textbf{Intermediate artifact edits}: downstream state must change while upstream evidence remains fixed.
\item \textbf{Cumulative revisions}: the workflow is revised repeatedly rather than discarded after one answer.
\item \textbf{Shared maintained artifacts}: humans or other agents inspect and revise intermediate outputs directly.
\item \textbf{Asymmetric recomputation cost}: some stages are expensive or slow enough that global reruns are operationally undesirable.
\end{itemize}

These properties describe a broad class of realistic AI-native work: research briefs, analysis pipelines, coding workflows, forecasting systems, compliance reviews, and operational runbooks. In such settings, the key question is rarely ``can the model produce an answer?'' The harder question is whether the system can evolve that answer over time without losing its own structure.

\subsection{What This Paper Does Not Claim}

The argument here should not be overstated. We do not claim that deterministic graphs replace all useful forms of agentic exploration. Open-ended discovery, broad brainstorming, and ambiguous early-stage problem framing may still benefit from unconstrained loops, reflection, and search. Nor do we claim that explicit execution structure by itself guarantees better one-shot answer quality. A poorly designed graph can still route bad prompts, weak tools, or misleading inputs.

What we claim is narrower and more systems-oriented. Once a workflow has repeated stages, reusable intermediate state, and nontrivial update pressure, leaving its structure implicit becomes a liability. At that point, the graph is not merely one possible representation among many. It becomes a practical mechanism for preserving correctness, reducing unnecessary recomputation, and making the system legible to its operators.

\subsection{Limitations}

This evaluation is intended as a controlled mechanism study rather than a comprehensive benchmark. It uses two controlled update tasks in one policy-memo domain, three system conditions, three repeats, and one model family. This scope is appropriate for isolating dependency isolation and artifact-level propagation, but it does not establish broad performance claims across all agentic workflows.

The loop baselines are naturalistic but not exhaustive; stronger custom loop harnesses with explicit state tracking could narrow some gaps. The DAG implementation is sequential, so wall-clock time does not always improve even when input context and recomputation are reduced. The graph must also be correctly specified: dependency mistakes, poor source routing, or lossy artifact interfaces can reduce the advantage. The experiments expose a mechanism by which partial cross-artifact inconsistency could lead to state drift, but they do not directly measure multi-round accumulation; longitudinal revision studies are needed to test that hypothesis. Finally, the results should not be read as evidence that DAG replay generally produces better prose than a holistic loop rewrite.

\section{Conclusion}

We introduced execution lineage as an execution substrate for AI-native work represented as artifact-producing DAGs with explicit dependencies. The empirical results sharpen the claim: execution lineage is not primarily a technique for making models better one-shot writers. Its value is in maintaining work under change.

The controlled update tasks show both sides of execution lineage. In the unrelated-branch update, DAG replay preserved stable work exactly and isolated an adjacent but non-dependent branch. In the intermediate-artifact edit, DAG replay propagated a meaningful change through downstream artifacts while preserving upstream state. Loop baselines were able to update visible final memos, but they were less reliable at preserving stable outputs and maintaining cross-artifact consistency.

These results suggest that persistent AI workflows need evaluation criteria beyond final answer quality. For long-lived work, correctness includes knowing what changed, what did not, and why. Execution lineage provides a systems abstraction for making those properties explicit.

\bibliographystyle{plain}

\end{document}